\def\year{2018}\relax
\documentclass[letterpaper]{article} 
\usepackage{aaai18}  
\usepackage{times}  
\usepackage{helvet}  
\usepackage{courier}  
\usepackage{url}  
\usepackage{graphicx}  
\usepackage{xcolor}
\usepackage{url}
\usepackage{amsmath,amscd,amsbsy,amssymb,latexsym,url,bm}

\begin{document}
	
	\title{Face Transfer with Generative Adversarial Network}
	\author{
		Runze Xu, Zhiming Zhou, Weinan Zhang, Yong Yu\\
		Shanghai Jiao Tong University
	}
	\maketitle
	\begin{abstract}
		Face transfer animates the facial performances of the character in the target video by a source actor. Traditional methods are typically based on face modeling. We propose an end-to-end face transfer method based on Generative Adversarial Network. Specifically, we leverage CycleGAN to generate the face image of the target character with the corresponding head pose and facial expression of the source. In order to improve the quality of generated videos, we adopt PatchGAN and explore the effect of different receptive field sizes on generated images.
		
	\end{abstract}
	
	\section{Introduction}
	Face transfer is a method for mapping face performances of one individual to facial animations of another one. It uses facial expressions and head poses from the video of a source actor to generate a video of a target character. A variety of methods have been developed for face transfer and have achieved impressive results. Previous work typically models the face of source and target, and then transfers the corresponding features from the source to the target, and finally re-renders the target face and blends to the original target image to achieve face transfer \cite{vlasic2005face,shi2014automatic,thies2016face2face}.  A data driven approach is proposed by \cite{li2012data}. They retrieved frames from a database based on a similarity metric and used optical flow as appearance and velocity measure and then searched for the $k$-nearest neighbors based on time stamps and flow distance. 

	Different from the methods mentioned above which divide the task into several steps, and explicitly model the facial attributes, in this paper, we use deep neural network to develop an end-to-end approach \cite{lecun2015deep}. Our work takes a talking video of a source actor as input. For every frame in the video, a face image of target character with corresponding facial expression and head pose is generated. By combining every generated frame, a corresponding video of the target character is generated.
	
	Face transfer is a special case of image-to-image translation tasks \cite{isola2016image,zhu2017unpaired}. The characteristic of the input video is the appearance of the character in the video, while the identity of each frame is the character's facial expression and head pose. Our method is to use Generative Adversarial Network (GAN) \cite{goodfellow2014generative} to learn to transform the characteristic while preserving the identity. There are two key factors for this task. First, a source face image should be mapped to a target face image with the corresponding facial expression and head pose. Second, the image should be of high quality, i.e., looks natural and indistinguishable to human.
	
	In this paper, to generate images with matching facial expression and head pose, we use CycleGAN \cite{zhu2017unpaired} to transfer the identity between two image sets. CycleGAN is proposed to capture the special characteristics of one image collection and translate the characteristics into the other image collection in the absence of any paired training samples. CycleGAN learns a one-to-one mapping, which ensures each input face image can be mapped to a target image with a corresponding facial expression and head pose \cite{kim2017learning}. 
	
	GAN is originally proposed to map the inputs to a real data distribution. A discriminator that models the whole images requires the generated images to be close to real images, which may restrict the creativity of the generator. \cite{isola2016image} proposed that PatchGAN is an effective architecture in image-to-image translation tasks. Instead of a discriminator that performs a judge from the whole image. PatchGAN discriminator has a receptive field smaller than the whole image. It only models images from patch level and explicitly requires every image patch to be real. In that case, the generated image, composed by realistic image patches, can be more diverse, which enhances the generator's creativity. We study the impact of different receptive field sizes on the generators and use a model with a pair of discriminators with a big and small receptive fields respectively to capture both global coherence and local patterns.
	
	
	To sum up, our contributions are as follows.
	\begin{itemize}
		\item To the best of our knowledge, our work is the first that applies Generative Adversarial Network to perform end-to-end face transfer which we formalize as an image-to-image translation problem.
		\item We explore the impact of discriminators with different receptive field sizes on the quality of generated images. We propose an architecture of two discriminators with different receptive field sizes. This enables the generator to create images with a head pose that does not occur in the real image set.
	\end{itemize}
	
	The demo video is provided at \url{goo.gl/RBbR9y}.
	
	\section{Related Work}
	\subsubsection{Face Transfer}
	\cite{vlasic2005face} performed face transfer based on a multi-linear model of 3D face meshes that separably parameterizes the space of geometric variations due to different facial attributes. The authors tracked a face template and re-rendered it under different expression parameters. \cite{dale2011video} tracked the facial performance in both videos. The authors warped the source to the target face and re-timed the source to match the target performance using the corresponding 3D geometry. \cite{garrido2014automatic} proposed a reenactment pipeline conceived as part image retrieval and part face transfer. \cite{li2012data} took advantage of an existing facial performance database of the target person. They used a query image to retrieve frames from a database based on similarity metrics. \cite{thies2016face2face} investigated face trackers and expression modeling to transfer facial expressions and achieved real-time face transfer. 
	
	Compared with above works, our approach uses generative adversarial network to achieve an end-to-end face transfer between two given characters without any supervision. We directly generate the target frames with the input frames of the source character.

	\subsubsection{Image-to-Image Translation}
	\cite{isola2016image} pointed out that many problems in image processing, computer graphics, and computer vision can be formulated as an image-to-image translation task. For example, label to scene, aerial to map, day to night, edges to photo and also grayscale to color. Some problems in face synthesizing can also be regarded as image-to-image translation tasks. To be specific, changing attributes of face images, such as gender, hair style \cite{kim2017unsupervised}, age, expression, beard and glasses \cite{shen2016learning}. In this paper, we also formulate face transfer as an image-to-image translation task.

	\subsubsection{Generative Adversarial Networks}
	Generative Adversarial Networks (GAN) \cite{goodfellow2014generative} has attained much attention in unsupervised learning during the recent 3 years. Conditional GAN, as a variant of GAN, is widely used in various computer vision scenarios. In some image-to-image translation tasks, the inputs are images rather than noises. \cite{zhu2017unpaired,kim2017unsupervised,yi2017dualgan} investigated similar cycle architecture and named this architecture as CycleGAN, DiscoGAN, DualGAN respectively. In this paper, we refer to this architecture as CycleGAN. 
	
	Compared with traditional GAN which has only a generator $G$ mapping domain $X$ to domain $Y$ and a discriminator on domain $Y$. CycleGAN adds another generator $F$ mapping domain $Y$ to domain $X$ and a discriminator on domain $X$. The two GANs form a cycle transformation, and cycle consistency loss is introduced to urge the cycle transformation to be identical. Such a cycle architecture can be applied to unpaired data \cite{zhu2017unpaired}. 
	We leverage cycle architecture to transfer the facial performance from the source character to the target character, with two unpaired videos, one for each character.

	\cite{isola2016image} proposed PatchGAN as an effective architecture in image-to-image translation tasks. It restricts the discriminator to image patches in order to model high-frequency structures. And the authors showed that PatchGAN is effective on a wider range of problems. A similar PatchGAN architecture was previously proposed in \cite{li2016precomputed}, for the purpose of capturing local style statistics. Such a discriminator models the image as a Markov random field \cite{li2016precomputed}. PatchGAN is also used in further studies such as unpaired settings \cite{zhu2017unpaired} and dual learning \cite{yi2017dualgan}. In this paper, we explore the effect of PatchGAN discriminators with different receptive field sizes.

	
	
	\subsubsection{GAN with Multi-Discriminators}
	Despite the impressive results achieved by GANs. GANs are reputably difficult to train. It is hard to balance the generator and the discriminator, and it easily gets mode-collapse. \cite{durugkar2016generative} extended GANs to multiple discriminators. For a generator $G$, $N$ discriminators of the same structure with random initialization are utilized as teachers for the generator. They suggested that GANs with multiple discriminators can better approximate the optimal discriminator, and are more stable on providing reliable feedback for the generator.
	
	In this paper, we experiment with the framework of two discriminators against one generator. Note that our discriminators are PatchGAN discriminators. We explore the case of two discriminators with the same structure and the case of two discriminators with different receptive field sizes. For the latter case, one discriminator has a large receptive field and another has a small receptive field. 
	
	\section{Preliminaries}
	In this section, we discuss some preliminaries of our models, including GANs and CycleGAN framework.
	
	A Generative Adversarial Network is a generative model that consists of two neural networks. A generator $G$ learns to map random noise vector $z$ to real data distribution: $ G : z \rightarrow x $. A discriminator $D$ tries to distinguish real data from generated samples. They are iteratively trained to play a two-player min-max game. 
	
	In image-to-image translation tasks, the generator takes in images as input instead of noise and maps images from $X$ to target domain $Y$: $ G : X \rightarrow \emph{Y} $. We follow the choice of \cite{zhu2017unpaired} which adopts least square loss instead of the negative log likelihood. The adversarial loss is defined as
	\begin{align}
	\emph{L}_{GAN}(G,D_Y) = &\mathbb{E}_{\emph{y} \sim P_{data}(y) }[(D_Y(y)-a)^2]  \nonumber\\ 
	&~~~ +\mathbb{E}_{\emph{x} \sim P_{data}(x) }[D_Y(G(x))^2].
	\end{align}
	
	CycleGAN, in addition to traditional GANs, adopts two pairs of generator and discriminator. A generator $G$ that maps $X$ to $Y$ with a discriminator on domain $Y$ and a generator $F$ that maps $Y$ to $X$ with a discriminator on domain $X$. The two GANs are trained simultaneously. Each image $x$ in domain $X$ transformed by $G$ to domain $Y$ is then transformed back to domain $X$ by $F: x \rightarrow G(x) \rightarrow F(G(x))$. CycleGAN introduces a cycle consistency loss which enforces $x$ and $G(F(x))$ to be consistent. Such a cycle architecture can thus be applied to unpaired data.
	
	The loss function of cycle consistency loss is defined as
	\begin{align} \label{org_cycle_loss}
	\emph{L}_{cyc}(G,F)=&\mathbb{E}_{\emph{x} \sim P_{data}(x)}[\lvert F(G(x))-x\rvert] \nonumber\\ 
	&~~~ +\mathbb{E}_{\emph{y} \sim P_{data}(y)}[\lvert G(F(y))-y \rvert],
	\end{align} 
	where we follow \cite{zhu2017unpaired,yi2017dualgan} to adopt L1 norm to measure the cycle consistency loss\footnote{We also experimented with L2 norm, but found that L1 norm is more robust to mode collapse.}. 
	
	The cycle consistency loss consists of two parts. They are conditioned on both $G$ and $F$. When minimizing $\mathbb{E}_{\emph{x} \sim P_{data}(x)}[\lvert F(G(x))-x)\rvert]$, \emph{G} is optimized to transform a real image $x$ to a generated sample that contains sufficient information to be transformed back to $x$. When minimizing $\mathbb{E}_{\emph{y} \sim P_{data}(y)}[\lvert G(F(y))-y \rvert]$, $G$ is optimized to adapt a fake sample $F(y)$ back to a real image $y$. 
	
	\section{Our Method}
	
	We empirically find that the original definition of the cycle consistency loss  makes the generators prone to generate artifacts meaningless to human. Thus we make a small modification: both generators only take real images as input and will not be trained on fake images during training:
	\begin{align}
	L_{cyc}(G) = \mathbb{E}_{\emph{x} \sim P_{data}(x)}[\lvert F(G(x))-x)\rvert] \nonumber\\
	L_{cyc}(F) = \mathbb{E}_{\emph{y} \sim P_{data}(y)}[\lvert G(F(y))-y)\rvert]
	\end{align}
	Such a modification helps reduce some artifacts.
	
	Original CycleGAN fixes the weight of adversarial losses to 1.0 and the two parts of cycle consistency share a weight parameter. We find different datasets are different in susceptibility to mode collapse. And the ratio of cycle loss and adversarial loss should be different for different datasets. Therefore, we introduce weights for different cycle passes as hyperparameters, and define the full objective function as
	\begin{align}
	L(G,F,D_X,D_Y) = &\alpha L_{GAN}(G,D_Y) + \beta L_{GAN}(F,D_X) \nonumber\\
	&~~~+ \lambda L_{cyc}(G)  + \lambda L_{cyc}(F),
	\end{align}
	where $\alpha$, $\beta$ and $\lambda$ are the hyperparameters.

	\subsection{PatchGAN Discriminators}
	For discriminators, we employ Markovian PatchGAN discriminator \cite{isola2016image,li2016precomputed},
	which models images only at patch level rather than the whole image. It assumes independence between pixels separated by more than a patch diameter. The discriminator is run convolutionally across the image and the losses of all image patches are averaged to provide the final loss of the discriminator. 
	PatchGAN discriminator is effective in capturing local high-frequency features but less effective in modeling global structure. 
	
	The choice of receptive field size is a staggering problem. In some image-to-image translation tasks, such as edge to photo, labels to street scene and grayscale to color \cite{isola2016image}, the generator changes the local style while preserving spatial information. The shape of the objects in the picture usually remains unchanged. The receptive field of discriminators in these tasks could be more arbitrary than that of our case. For example, in \cite{isola2016image}, they experimented $1\times1$, $16\times16$, $70\times70$, $256\times256$ on $256\times256$ images and got visually similar results with $256\times256$ ImageGAN and $70\times70$ PatchGAN.
	
	In our face transfer task, we find the receptive field size strongly affects the quality of generated faces. With a small receptive field, only generated image patches are required to be realistic, which results in more diverse generated images and enhances the generator's creativity, especially on a dataset with limited samples. While, for the fact that we are modeling an image of a entire face, we need a discriminator with a receptive field close to image size. If the receptive field is too small, it will result in unreasonable deformation of generated faces.  
	
	\subsection{Multiple Discriminators}
	A discriminator with limited capacity may fail to generate realistic images. However, modification on the discriminator's structure is often along with a difficulty in training GANs. When the discriminator reaches a far superior situation to the generator, the generator may stop making progress \cite{durugkar2016generative,arjovsky2017towards,neyshabur2017stabilizing}. When the distribution of generated samples has little overlap with real image distribution, the generator cannot receive efficient gradients from the discriminator to improve its performance, which is referred as the gradient vanishing problem in GAN \cite{arjovsky2017wasserstein}.
	
	We introduce the scheme of training two discriminators against one generator as a more stable way to improve the capacity of discriminators, as similar to \cite{durugkar2016generative}. To be specific, the multi-discriminator architecture can better approximate the optimal discriminator, and, if one of the discriminators is trained to be far superior over the generator, the generator can still receive instructive gradients from the other discriminator. 
	
	\subsection{Choice of Receptive Fields}
	In the two-discriminator setting, we conduct experiments with three different pairs of receptive field sizes. To ensure justice, all models have two discriminators against one generator. The performance of these models will be shown and analyzed in Experiments Section. 
	
	For the first case, both discriminators are with  receptive field sizes of the size $97\times97$. The two discriminators share the same structure but are randomly instantiated. We simply average their losses as the final adversarial loss.  Such a receptive field is close to image size. It models real images from a global view. Therefore, if a generated face image exhibits a pose that never occurs in the real image set, the discriminator will give a low score and prevent the generator from generating such images, which restricts the creativity of the generator.
	
	For the second case, we use two discriminators with $42\times42$ receptive fields. Because the $42\times42$ discriminator only models local patterns of real images and it does not require the whole generated image to be similar to real images. The generator has little restriction on the global structure, which enables the generator to transform images with head poses that never occur in the target domain into a much better sample than the case mentioned above. In other words, it enhances the generator's creativity. However, discriminators with small receptive fields cannot model global features, which results in global inconformities such as excessive deformation of generated eyes and face.  
	
	For the third case, one discriminator has a $97\times97$ receptive field, while the other has a $42\times42$ receptive field. It is a trade-off between the two cases mentioned above that models global and local structure simultaneously. We add parameters to tune weight of the two adversarial losses. We hope this model can take both global features and local features into consideration. While the $42\times42$ discriminator improves the generator's creativity, the $97\times97$ discriminator can help eliminate those abrupt deformation caused by the absence of global inspection.
	
	For the mapping function $ \emph{G} : \emph{X} \rightarrow \emph{Y} $, The formulation of the final adversarial loss is defined as
	\begin{align}
	L_{GAN}(G,D_{Y_1},D_{Y_2}) =& \gamma \cdot L_{GAN}(G,D_{Y_1})\nonumber\\&+(1-\gamma)\cdot L_{GAN}(G,D_{Y_2}),
	\end{align}
	where $\gamma$ is the hyperparameter to maintain the ratio between the adversarial losses of the two discriminators.
	Note that $D_{Y_1}$ and $D_{Y_2}$ are two discriminator instances. In the first two cases, $\gamma$ is set as 0.5 as the balance of the two factors are not sensitive to the final results. In the third case, $\gamma$ needs to be carefully tuned as it is sensitive to the performance.
	
	\section{Experiments} \label{Experiments}
	\subsection{Datasets}
	We conduct experiments on three clips:
	\begin{itemize}
		\item Barack Obama in weekly address
		\item Joe Biden in weekly address
		\item Li Xiuping in CCTV News
	\end{itemize}
	
	We manually crop the 3 videos and extract the frames of the cropped video. All of these video clips are mainly a character talking to the camera. We choose these videos based on the following reasons. Because we aim at transferring the facial performance, each video includes the facial part of a acting character. In our video clips, body parts usually do not appear in the cropped video and the shooting angles are fixed, otherwise, irrespective body movements and the changing of shooting angle may also be considered as important identities by the generator. 
	The images are scaled to $128\times128$. 
	
\begin{figure*}[t]
		\centering
		\vspace{0pt}
		\includegraphics[width=2.1\columnwidth]{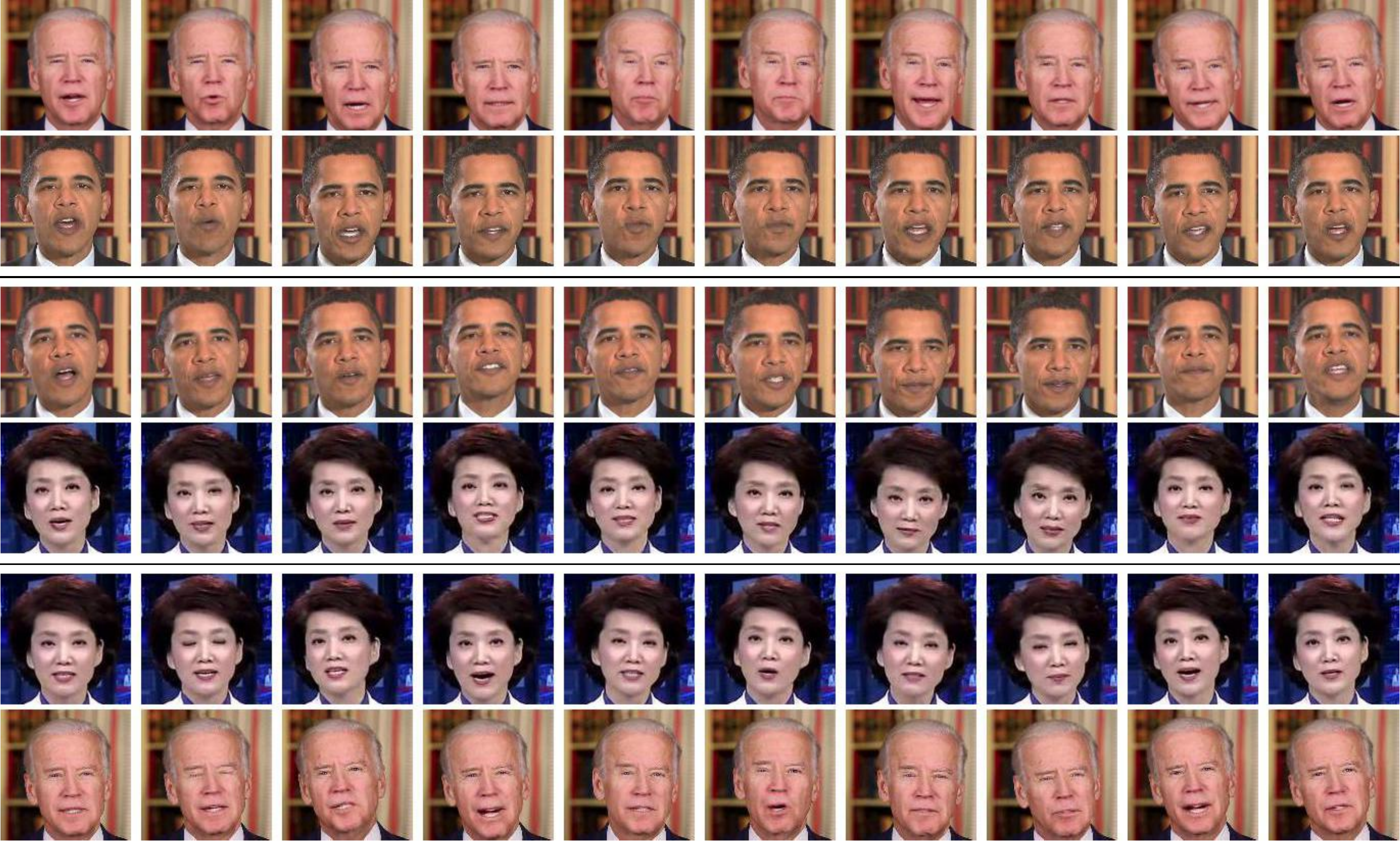}\\
		\caption{
			Example results: the odd rows are real source images and the even rows are generated target images.} \label{fig:example}
		\vspace{0pt}	
	\end{figure*}

	Most input images with common facial expression and head pose are transformed into realistic samples in the target domain. Example results are shown in Figure \ref{fig:example}.
	
	
	
	\subsection{Performance of Different Receptive Fields}
	In this section, we compare the performance of three models. They all have two discriminators trained against one generator. The first model has two PatchGAN discriminators with $97\times97$ receptive field, represented by $97+97$ model. The second has two discriminators with $42\times42$ receptive field, represented by $42+42$ model.The third has a discriminator with $97\times97$ receptive field and a discriminator with $42\times42$ receptive field, represented by $97+42$ model. 
	
	To illustrate the difference between the three models, we compare their performance on the task \emph{Joe Biden $\rightarrow$ Li Xiuping}. This is the most difficult task in the three datasets for two reasons. First, Joe Biden is an English speaker while Li Xiuping is a Chinese speaker. There is no counterpart for some mouth shapes in Joe Biden's video. Second, Joe Biden moves his head arbitrarily in his video while Li Xiuping moves in a small range. We pick the representative frames that illustrate the advantages and disadvantages of the three models. 
	
	$97+97$ discriminator models global structure of images, which restricts the generator's creativity. It makes the generator generates noisy and distorted face images if the source image exhibits a head pose unseen in target image set. 
	$42+42$ discriminator allows the generator to generate uncommon head pose but cannot ensure global features. The $97+42$ model is a trade-off between global and local features. By tuning the weight of two discriminators' adversarial losses, we achieved high-quality results. 
	
	\subsection{Creativity on Unseen Head Poses}
	In the video of Joe Biden, he sometimes raises his head, sometimes tilts his head drastically,  while Li Xiuping never does so in her video. So there is no image that can be considered as a direct reference. 
	
\begin{figure*}  
		\vspace{-0pt}
		\begin{minipage}[t]{0.5\linewidth}  
			\centering
			\vspace{2pt}
			\includegraphics[width=1.0\columnwidth]{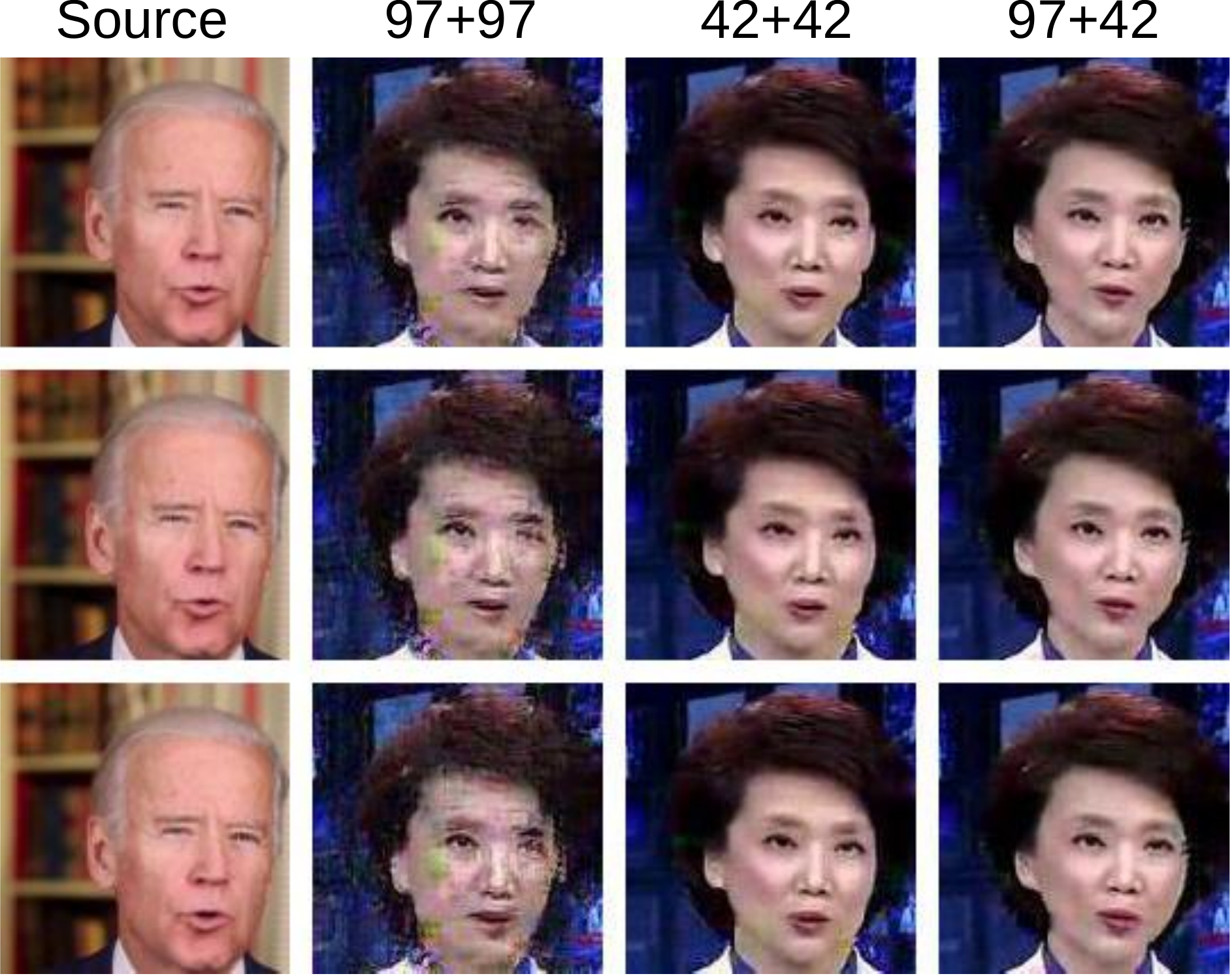}\\
			\caption{The first column are three frames in succession. $97+97$ model fails to generate clear images. $42+42$ model generates much better images but results in abrupt deformation. $97+42$ model alleviates the deformation and achieves images close to real images.} 
			\label{fig:deformation}
		\end{minipage}%
		\hspace{10pt}
		\begin{minipage}[t]{0.5\linewidth}  
			\centering
			\vspace{1.6pt}
			\includegraphics[width=1.0\columnwidth]{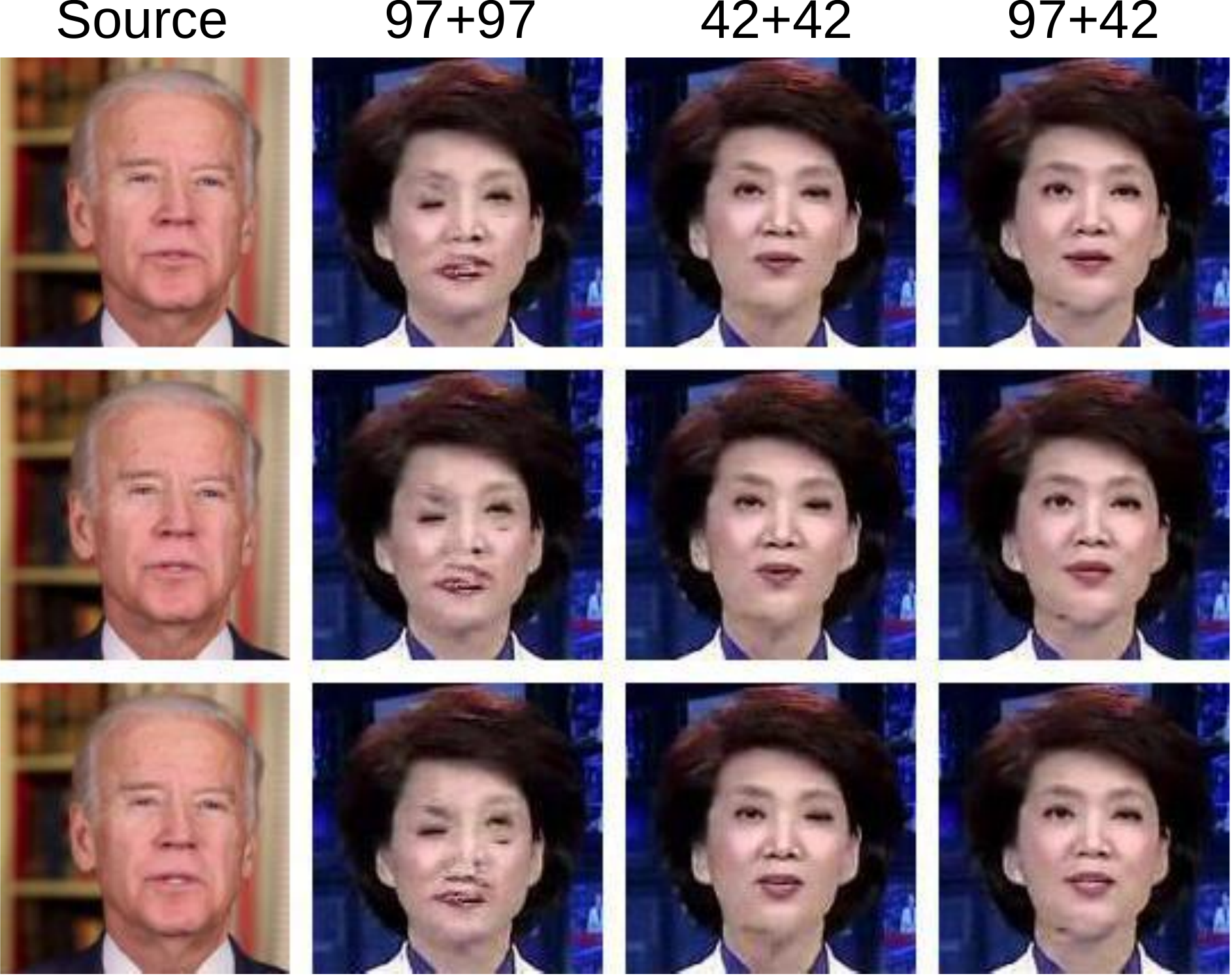}\\
			\caption{The first column are three source frames in succession. The other columns are results of different models. $97+97$ model generates distorted faces. $42+42$ and $97+42$ model generate similar result. } \label{fig:raise}
			\vspace{0pt}
		\end{minipage}  
		\vspace{0pt}
	\end{figure*}

	Joe Biden in the source frames raises his head, as shown in Figure \ref{fig:raise}. The $97+97$ model generates distorted faces, for the discriminator models the whole images and it considers the unseen head pose as a fake sample, which hinders the generator from creating unseen head poses. The other two models generate meaningful faces, although there are some artifacts on the neck. The results of $42+42$ model and $97+42$ model are similar.
	
	In another video fragment, Figure \ref{fig:deformation}, Joe Biden tilts his head drastically. $97+97$ model generates a face image with noise everywhere. $42+42$ model generates a clear face, but causes unpleasant deformation of the generated face. This is because the $42+42$ discriminator cannot model the whole face. It only inspects generated images from a local perspective. Without the restriction on global features, it cannot penalize the obvious global deformation. $97+42$ discriminator models both global and local structures, which leads to much better results.
	
	\subsection{Limitation}
	
\begin{figure}[t]
		\centering
		\vspace{3.5pt}
		\includegraphics[width=1.0\columnwidth]{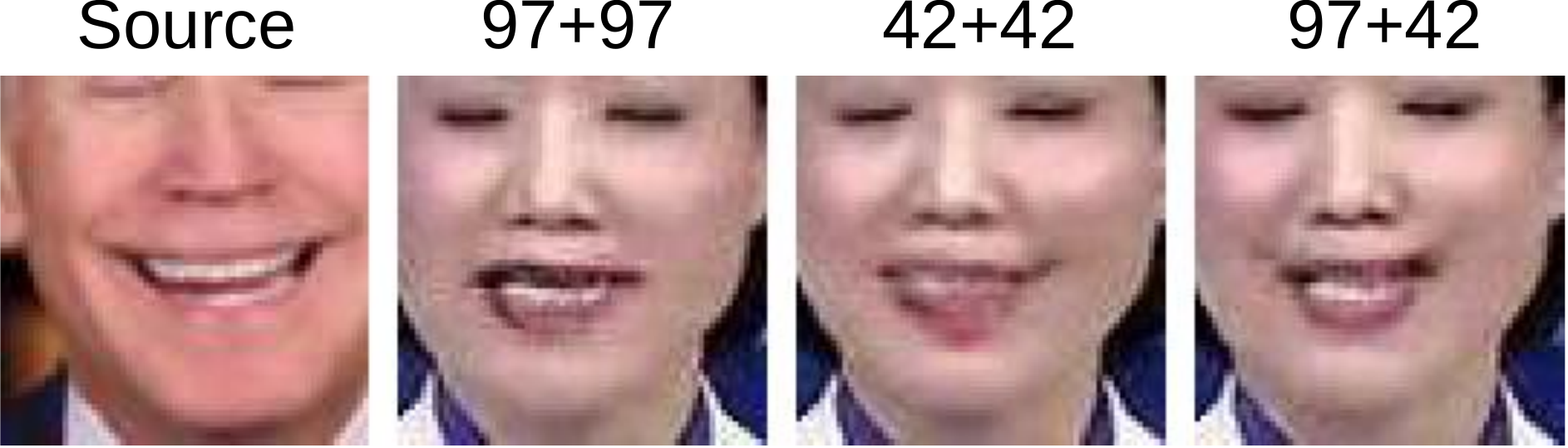}\\
		\caption{The performance of three models on generating an unseen mouth. } \label{fig:mouth}
		\vspace{0pt}
	\end{figure}

	In this section, we discuss the limitation of our face transfer method. We show the performance of our three models on generating unseen expressions in Figure \ref{fig:mouth}. We cut a fragment of video in which Joe Biden grins broadly to pronounce 'any'. It is hard to find a similar mouth shape in Chinese, let alone a short video of Li Xiuping. Without a reference, it is hard to generate a sharp and realistic mouth with our approach. $97+42$ model generates a mouth that resembles the source image in the overall shape but cannot create the details of teeth. 
	
	CycleGAN encourages the generator to map a source image to a target image with identical facial expression and head pose. However, when the source and target video do not match in the diversity of head pose and facial expressions, it is hard to learn a perfect one-to-one mapping, which results in noisy and distorted results as shown in Figure 2, Figure 3 and Figure 4. To address this issue we adopt our two-discriminator architecture to create some unseen poses. 
	
	
	
	
	\section{Conclusions}
	In this paper, we leverage CycleGAN and PatchGAN to achieve an end-to-end face transfer. CycleGAN learns a one-to-one mapping, which ensures each source face image to be mapped to a target image with a corresponding facial expression and head pose.
	Practically, with limited training samples, it is difficult to well generate the corresponding target face image with unseen facial expression or head pose in the target dataset. To improve the generalization ability of the generator, we propose to adopt a discriminator with a small receptive field to alleviate the restriction on the generator and a discriminator with a big receptive field to ensure global coherence. This two-discriminator architecture achieves the best result in our experiments.
	
	For the future work, loss re-weighting on image patches could help improve generated image quality. And an investigation on the impact of receptive field size on other image-to-image translation tasks is also interesting.
	
	
	\bibliographystyle{aaai}
	\bibliography{face_transfer.bib}

\end{document}